\documentclass[11pt]{article}
\usepackage{acl-hlt2011}
\usepackage{times}
\usepackage{latexsym}
\usepackage{amsmath}
\usepackage{amsthm}
\usepackage{amsfonts}
\usepackage{multirow}
\usepackage{url}
\usepackage{algorithmic}[1]
\usepackage{graphicx}
\DeclareMathOperator*{\argmax}{arg\,max}
\title{Large-Margin Learning of Submodular Summarization Methods}

\author{Ruben Sipos \\
  Dept. of Computer Science \\
  Cornell University \\
  Ithaca, NY 14853 USA \\
  {\tt rs@cs.cornell.edu} \\\And
  Pannaga Shivaswamy \\
  Dept. of Computer Science \\
  Cornell University \\
  Ithaca, NY 14853 USA \\
  {\tt pannaga@cs.cornell.edu} \\\And
  Thorsten Joachims \\
  Dept. of Computer Science \\
  Cornell University \\
  Ithaca, NY 14853 USA \\
  {\tt tj@cs.cornell.edu} \\
}

\date{}
\newcommand{\argmin}{\operatornamewithlimits{argmin}}
\newtheorem{definition}{Definition}

\newcommand{\w}{{\bf w}}  

\begin{document}
\maketitle


\begin{abstract}
In this paper, we present a supervised learning approach to training submodular scoring functions for extractive multi-document summarization. By taking a structured predicition approach, we provide a large-margin method that directly optimizes a convex relaxation of the desired performance measure. The learning method applies to all submodular summarization methods, and we demonstrate its effectiveness for both pairwise as well as coverage-based scoring functions on multiple datasets. Compared to state-of-the-art functions that were tuned manually, our method significantly improves performance and enables high-fidelity models with numbers of parameters well beyond what could reasonbly be tuned by hand.
\end{abstract}


\section{Introduction}

Automatic document summarization is the problem of constructing a short text describing the main points in a (set of) document(s). 
Example applications range from generating short summaries of news articles, to presenting snippets for URLs in web-search. 
In this paper we focus on extractive multi-document summarization, where the final summary is a subset of the sentences from multiple input documents. In this way, extractive summarization avoids the hard problem of generating well-formed natural-language sentences, since only existing sentences from the input documents are used.

A current state-of-the-art method for document summarization was recently proposed by Lin and Bilmes \cite{Lin2010}, using a submodular scoring function based
on inter-sentence similarity. On the one hand, this scoring function rewards summaries that are similar to many sentences in the original documents (i.e. promotes coverage). On the other hand, it penalizes summaries that contain sentences that are similar to each other (i.e. discourages redundancy). While obtaining the exact summary that optimizes the objective is computationally hard, they show that a greedy algorithm is guaranteed to compute a good approximation. However, their work does not address how to select a good inter-sentence similarity measure, leaving this problem as well as selecting an appropriate trade-off between coverage and redundancy to manual tuning. 

To overcome this problem, we propose a supervised learning method that can learn both the similarity measure as well as the coverage/reduncancy trade-off from training data. Furthermore, our learning algorithm is not limited to the model of Lin and Bilmes \cite{Lin2010}, but applies to all submodular summarization models. Due to the diminishing-returns property of submodular set functions and their computational tractability, this class of functions provides rich space for designing summarization methods. To illustrate this point, we also provide experiments for a submodular coverage-based model originally developed for diversified information retrieval \cite{EssentialPages}.

In general, our method learns a parameterized submodular scoring function from supervised training data, and its implementation is available for download\footnote{\url{http://www.cs.cornell.edu/~rs/sfour/}}. Given a set of documents and their summaries as training examples, we formulate the learning problem as a structured prediction problem and derive a maximum-margin algorithm in the structural SVM framework. Note that, unlike other learning approaches, our method does not require a heuristic decomposition of the learning task into binary classification problems \cite{r17}, but directly optimizes a structured prediction. This enables our algorithm to directly optimize the desired performance measure (e.g. ROUGE) during training. Furthermore, our method is not limited to linear-chain dependencies like \cite{r5,r26}, but can learn any submodular scoring function.

This ability to easily train summarization models makes it possible to efficiently tune models to various types of document collections. In particular, we find that our learning method can reliably tune models with 
hundreds of parameters based on a training set of 
about 30 examples. This increases the fidelity of models compared to their hand-tuned counterparts, showing significantly improved empirical performance. We provide a detailed investigation into the sources of these improvements, identifying further directions for research.


\section{Related work}

Work on extractive summarization spans a large range of approaches. 
Starting with unsupervised methods,
one of the widely known approaches is MMR \cite{r3}. It uses a greedy 
approach for selection and considers the trade-off between relevance 
and redundancy. Later it was extended \cite{r9} to support multi-document 
settings by incorporating additional information available in this 
case. Good results can be achieved by reformulating this as a 
knapsack packing problem and solving it using dynamic programing \cite{McD07}.

A popular stohastic graph-based summarization method is LexRank \cite{Lex04}. 
It computes sentence importance based on the concept of eigenvector 
centrality in a graph of sentence similarities. Similarly, 
TextRank \cite{r21} is also graph based ranking system for identification 
of important sentences in a document by using sentence similarity 
and PageRank \cite{r2}. Sentence extraction can also be implemented using 
other graph based scoring approaches \cite{Mih04} such as HITS \cite{r15} and 
positional power functions. Graph based methods can also be paired 
with clustering such as in CollabSum \cite{r30}. This approach first uses 
clustering to obtain document clusters and then uses graph based 
algorithm for sentence selection which includes inter and intra-document 
sentence similarities. Another clustering based algorithm \cite{r23} is 
diversity based extension of MMR that finds diversity by clustering 
and then proceeds to reduce redundancy by selecting a representative 
for each cluster.

The manually tuned sentence pairwise model \cite{Lin2010,LinACL} we took inspiration 
from is based on budgeted submodular optimization. A summary is produced 
by maximizing an objective function that includes coverage and 
redundancy terms. Coverage is defined as the sum of sentence similarities 
between the selected summary and the rest of the sentences, while redundancy 
is the sum of pairwise intra-summary sentence similarities. Another approach based 
on submodularity \cite{Qa} is relying on extracting important keyphrases 
from citation sentences for a given paper and using them to build the summary.

In the supervised setting, a lot of early methods \cite{r17} made 
independent binary decisions whether to include a particular sentence 
in the summary or not. This ignores dependencies between sentences 
and can result in high redundancy. The same problem arises when 
using learning to rank approaches such as ranking support vector 
machines, support vector regression and gradient boosted decision 
trees to select the most relevant sentences for the summary \cite{r19}.

Introducing some dependencies can improve the performance. One limited way of introducing dependencies between sentences is by using a linear-chain HMM. The HMM is assumed to produce the 
summary by having a chain transitioning between summarization 
and non-summarization states \cite{r5} while traversing the sentences 
in a document. A more expressive approach is using a CRF for sequence 
labeling \cite{r26} which can utilize larger and not necessarily independent feature spaces. The disadvantage of using linear chain models, however,
is that they represent the summary as a sequence of sentences. Dependencies between sentences that are far away from each other cannot be modeled efficiently.
In contrast to such linear chain models, our approach on submodular scoring functions can model long-range dependencies. In this way our method can use properties of
the whole summary when deciding which sentences to include in it.

More closely related to our work is that of \cite{Li}. They use the diversified retrieval method proposed in \cite{r31} for document summarization. Moreover, they assume that subtopic labels are available so that additional constraints for diversity, coverage and balance can be added to the structural SVM learning problem.
In contrast, our approach does not require
the knowledge of subtopics (thus allowing us to apply it to a wider range of 
tasks) and avoids adding additional constraints (simplifying the algorithm).
Furthermore, it can use different submodular objective functions, for example
word coverage and sentence pairwise models described later in this paper.

Another closely related work \cite{joinec} also takes learning approach
in the structural SVM framework to summarize a set of documents. However, they do not consider submodular functions, but instead solve an Integer Linear Program (ILP) or an approximation thereof. The ILP encodes a compression model where arbitrary parts of the parse 
trees of sentences in the summary can be cut and removed. This allows them to
select parts of sentences and yet preserve some
 gramatical structure. Their work focuses on learning a particular compression model, while our work explores learning a general and large class of sentence selection models.

\section{Submodular document summarization}
In this section, we illustrate how document summarization can be addressed using submodular set functions. The set of documents to be summarized is split into a set of individual sentences $x = \{s_{1},...,s_{n}\}$. The summarization method then selects a subset $\hat{y} \subseteq x$ of sentences that maximizes a given scoring function $F_x: 2^x \rightarrow \mathbb{R}$ subject to a budget constraint (e.g. less than $B$ characters). 
\begin{equation}
\label{eq:maxsum}
\hat{y} = \argmax_{y \subseteq x} F_x(y) \hspace{1cm}s.t. \:|y| \le B
\end{equation}
In the following we restrict the admissible scoring functions $F$ to be submodular.
\begin{definition}
Given a set $x$, a function $F:2^{x} \rightarrow \mathbb{R}$ is submodular
iff for all $u \in U$ and all sets $s$ and $t$ such that $s \subseteq t \subseteq x$,
we have, $$F(s \cup \{  u \}) - F(s) \ge F( t \cup \{ u \}) - F(t).$$
\end{definition}
Intuitively, this definition says that adding $u$ to a subset $s$ of $t$ increases $f$ at least as much as adding it to $t$. Using two specific submodular functions as examples, the following sections illustrate how this diminishing returns property naturally reflects the trade-off between maximizing coverage while minimizing redundancy.

\subsection{Pairwise scoring function}

\begin{figure}[h!]
\begin{center}
\includegraphics[width=\linewidth]{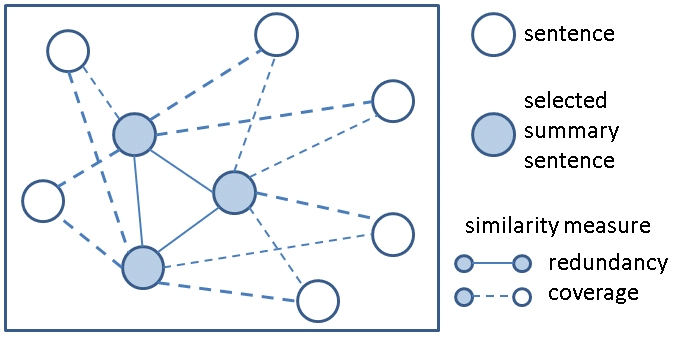}
\end{center}
\caption{
Illustration of the pairwise model. Not all edges are shown for clarity purposes.
Edge thickness denotes the similarity score.}
\label{fig:illupair}
\end{figure}

The first submodular scoring function we consider was proposed by \cite{Lin2010} based on a model of pairwise sentence similarities. It scores a summary $y$ using the following function, which \cite{Lin2010} shows is submodular.
\begin{equation}
\label{eq:submod1}
F_x(y) = \!\!\!\!\! \sum_{i \in x \backslash y, j \in y}\!\!\!{\sigma(i,j)}  - \lambda \!\!\sum_{i,j \in y : i \neq j}\!{\sigma(i,j)}.
\end{equation}
$\sigma(i,j) \ge 0$ denotes a measure of similarity between pairs of sentences $i$ and $j$. The first term in Eq.~\ref{eq:submod1} is a measure of how similar the sentences included in summary $y$ are to the other sentences in $x$. The second term penalizes $y$ by how similar its sentences are to each other. $\lambda > 0$ is a scalar parameter that trades off between the two terms.
Maximizing $F_x(y)$ amounts to increasing the similarity of the summary 
to excluded sentences while minimizing repetitions in the summary. An example is illustrated in Figure \ref{fig:illupair}. In the simplest case, $\sigma(i,j)$ may be the TFIDF \cite{Salton} cosine similarity, but we will show later how to learn sophisticated similariy functions.

\subsection{Coverage scoring function}

A second scoring function we consider was first proposed for diversified document retrieval \cite{r31}, 
but it naturally applies to document summarization as well \cite{Li}. It is based on a notion of word coverage, where each word $v$ has some importance weight $\omega(v) \ge 0$. A summary $y$ covers a word if at least one of its sentences contains the word. The score of a summary is then simply the sum of the word weights its covers (though we could also include a concave discount function that rewards covering a word multiple times \cite{Raman/etal/11a}). 
\begin{equation}
\label{eq:submod2}
 F_x(y) = \sum_{v \in V(y)} \omega(v)
\end{equation}
$V(y)$ denotes the union of all words in $y$. This function is analogous to a maximum coverage problem, which is known to be submodular \cite{maxcoverage}.

\begin{figure}[h]
\begin{center}
\includegraphics[width=\linewidth]{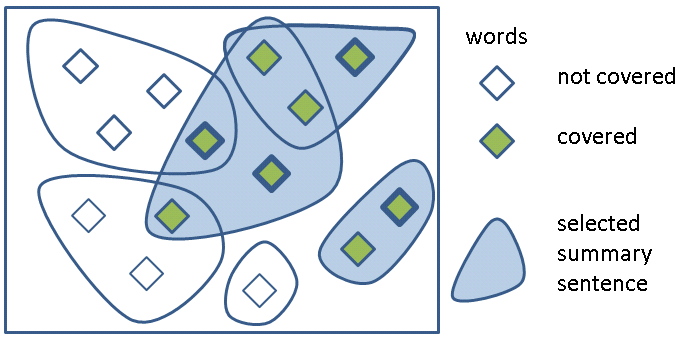}
\end{center}
\caption{
Illustration of the coverage model. Word border thickness represents importance.}
\label{fig:illucov}
\end{figure}

An example of how a summary is scored is illustrated in the Figure \ref{fig:illucov}. Analogous to the definition of similarity $\sigma(i,j)$ in the pairwise model, the choice of the word importance function $\omega(v)$ is crucial in the coverage model. A simple heuristic is to weigh words highly that occur in many sentences of $x$, but in few other documents \cite{EssentialPages}. However, we will show in the following how to learn $\omega(v)$ from training data.

\begin{figure}
\begin{center}
\hrule
\smallskip
\begin{algorithmic}
\STATE $\hat{y} \gets \emptyset$
\STATE $A \gets x$
\WHILE{$A \neq \emptyset$}
  \STATE $\displaystyle k \gets \argmax_{l \in A} \frac{F_x(\hat{y} \cup \{l\})-F_x(\hat{y})}{(c_{l})^{r}}$
  \IF{$c_k \!+\! \sum_{i \in \hat{y}}{c_i} \!\leq\! B$ \mbox{\bf and} $F_x(\hat{y} \cup \{\!k\!\})-F_x(\hat{y}) \!\ge\! 0$}
    \STATE $\hat{y} \gets \hat{y} \cup \{k\}$
   \ENDIF       
  \STATE $A \gets A \backslash \{k\}$
\ENDWHILE
\end{algorithmic}
\smallskip
\hrule
\end{center}
\caption{
Greedy algorithm for finding the best summary $\hat{y}$ given a
scoring function $F_x(y)$. Values $c_{i}$ represent costs of sentences (i.e. lengths).}
\label{fig:greedyalg}
\end{figure}

\subsection{Computing a Summary}

Computing the summary that maximizes either of the two scoring functions from above (i.e. Eqns. (\ref{eq:submod1}) and (\ref{eq:submod2})) is NP-hard \cite{McD07}. However, it is known that the greedy algorithm shown in Figure \ref{fig:greedyalg} can achieve a $1-1/e$ approximation to the optimum solution for any linear budget constraint \cite{Lin2010,maxcoverage}. Even further, this algorithm provides a $1-1/e$ approximation for any monotone submodular scoring function. 

The algorithm starts with an empty summarization. In each step, a
sentence is added to the summary that results in the maximum relative
increase of the objective. The increase is relative to the amount of
budget that is used by the added sentence. The algorithm terminates
when the budget $B$ is reached.

Note that the algorithm has a parameter $r$ in the denominator of the
selection rule, which \cite{Lin2010} report to have some impact on
performance. Selecting $r$ to be less than $1$ gives more importance
to ''information density'' (i.e. sentences that have a higher ratio of
score increase per length). The $1-\frac{1}{e}$ greedy approximation
guarantee holds despite this additional parameter \cite{Lin2010}.
More details on our choice of $r$ and its effects are provided in the
experiments section.


\section{Learning algorithm}
In this section, we propose a supervised learning method for training a submodular scoring function to produce desirable summaries. In particular, for the pairwise and the coverage model, we show how to learn the similarity function $\sigma(i,j)$ and the word importance weights $\omega(v)$ respectively. In particular, we parameterize $\sigma(i,j)$ and $\omega(v)$ using a linear model, allowing that each depends on the full set of input sentences $x$.
\begin{equation}
\label{eq:linpar}
\sigma_x(i,j) = \w^T \phi^p_x(i,j) \hspace{0.5cm} \omega_x(v) = \w^T \phi^c_x(v)
\end{equation}
$\w$ is a weight vector that is learned, and $\phi^p_x(i,j)$ and $\phi^c_x(v)$ are feature vectors. In the pairwise model, $\phi^p_x(i,j)$ may include feature like  the TFIDF cosine between $i$ and $j$ or the number of words from the document titles that $i$ and $j$ share etc. In the coverage model, $\phi^c_x(v)$ may include features like indicator of whether $v$ occurs in more than 10\% of the sentences in $x$ or whether $v$ occurs in the document title etc.

We propose to learn the weights following a large-margin framework using structural SVMs. Structural SVMs learn a discriminant function 
\begin{equation}
\label{eq:svmpred}
h(x) =  \argmax_{ y \in {\cal Y} } {\bf w}^\top \Psi(x,y)
\end{equation}
that predicts a structured output $y$ given a (possibly also structured) input $x$. $\Psi(x,y) \in \mathbb{R}^N$ is called the joint feature-map between input $x$ and output $y$. Note that both submodular scoring function in Eqns. (\ref{eq:submod1}) and (\ref{eq:submod2}) can be brought into the form $\w^T \Psi(x,y)$ for the linear parametrization in Eq.~(\ref{eq:linparpsi}) and (\ref{eq:linparpsi2}). 
\begin{align}
\label{eq:linparpsi}
\Psi^p(x,y)\!\!\!&= \!\!\!\!\!\!\!\!\!\sum_{i \in x \backslash y, j \in y}\!\!\!\!\!\phi^p_x(i,j)  - \lambda \!\!\!\sum_{i,j \in y : i \neq j}\!\!\!\phi^p_x(i,j)\\
\label{eq:linparpsi2}
\Psi^c(x,y)\!\!&= \!\!\!\!\!\sum_{v \in V(y)} \phi^c_x(v)
\end{align}
After this transformation, it is easy to see that computing the maximizing summary in Eq.~(\ref{eq:maxsum}) and the structural SVM prediction rule in Eq.~(\ref{eq:svmpred}) are equivalent.

To learn the weight vector $\w$, structural SVMs require training examples $(x^1,y^1), ..., (x^n,y^n)$ of input/output pairs. In document summarization, however, the ``correct'' extractive summary is typically not known. Instead, training documents $x^i$ are typically annotated with multiple manual (non-extractive) summaries (denoted by $Y^i$). To determine a single extractive target summary $y^i$ for training, we find the extractive summary that (approximately) optimizes ROUGE score -- or some other loss function $\Delta(Y^i,y)$ -- with respect to $Y^i$.
\begin{equation}
\label{eq:gold}
y_i = \argmin_{ y \in {\cal Y}} \Delta(Y^i,y) 
\end{equation}
We call the $y^i$ determined in this way the ``target'' summary for $x^i$.

\begin{figure}
\begin{center}
\hrule
\smallskip

\begin{algorithmic}
\STATE $\forall i : \mathcal{W}_{i} \gets \emptyset$
\REPEAT
  \FOR{$\forall i$} 
    \STATE $\displaystyle \hat{y} \gets \argmax_{y} w^{T} \Psi(x^{i},y) + \Delta(Y^{i},y)$
    \IF{$\displaystyle w^{T} \Psi(x^{i},y^{i}) + \epsilon \le w^{T} \Psi(x^{i},\hat{y}) + \Delta(Y^{i},\hat{y})-\xi_{i}$}
      \STATE $\mathcal{W}_{i} \gets \mathcal{W}_{i} \cup \{\hat{y}\}$
      \STATE $w \gets$ solve QP using constraints $\mathcal{W}_{i}$
    \ENDIF
  \ENDFOR
\UNTIL{no $\mathcal{W}_{i}$ has changed during iteration}
\end{algorithmic}

\smallskip
\hrule
\end{center}
\caption{
Cutting-plane algorithm for solving the learning optimization problem
using only polynomial number of steps to achieve a requested tolerance $\epsilon$. }
\label{fig:cpalg}
\end{figure}

Following the structural SVM approach, we can now formulate the problem of learning $\w$ as the following quadratic program (QP): 
\begin{align}
\label{eq:ssvm}
\min_{ {\bf w}, \xi \ge 0 } &\frac{1}{2} \| {\bf w } \|^2 + \frac{C}{n} \sum_{i=1}^n \xi_i \\
\text{s.t.~} & {\bf w}^\top \Psi(x^i,y^i) - {\bf w}^\top \Psi(x^i,\hat{y}^i) \ge \nonumber \\ \nonumber 
& \Delta(\hat{y}^i,Y^i) - \xi_i,~~ \forall \hat{y}^i \neq y^i,~~ \forall 1 \le i \le n .
\end{align}
The above formulation ensures that the scoring function with the target summary (i.e.
${\bf w}^\top \Psi(x^i,y^i)$) is larger than the scoring function for any other summary  $\hat{y}^i$
 (i.e., ${\bf w}^\top \Psi(x^i,\hat{y}^i)$). The objective function learns a large margin weight vector 
${\bf w}$ while trading it off with an upper bound on the empirical loss. The two quantities are traded off 
with a parameter $C > 0$. 

Even though the QP has exponentially many constraints in the number of sentences in the input documents, it can be solved in polynomial time via 
a cutting plane algorithm \cite{tso}.  The steps of the algorithm are shown in Figure \ref{fig:cpalg}.
In each iteration of the algorithm, for each training document $x^i$, a summary  $\hat{y}^i$ which
worst violates the constraint in \eqref{eq:ssvm} is found. This is done by solving
$$\displaystyle \hat{y} \gets \argmax_{y \in {\cal Y}} w^{T} \Psi(x^i,y) + \Delta(Y^i,y)$$
which can be done efficiently by the greedy algorithm in Figure~\ref{fig:greedyalg}.
After the worst violating constraint for each training example is added, the resulting quadratic program
is solved. These steps are repeated until all the constraints are satisfied to a required precision $\epsilon$.

Finally, special care has to be taken to appropriately define the loss function $\Delta$ given the disparity of $Y^i$ and $y^i$. Therefore, we first define an intermediate loss function as follows:
$$\Delta_{R}(Y,\hat{y}) = \max (0, 1 - ROUGE1_{F}(Y,\hat{y})),$$
based on the (slightly simplified) ROUGE-1 F score which is a standard metric for measuring
the quality of a document summarization. To ensure that the loss function
is zero for the target label as defined in \eqref{eq:gold}, we normalized the above
loss as below: 
$$\Delta(Y^i, \hat{y}) = \max (0, \Delta_{R}(Y^i,\hat{y}) - \Delta_{R}(Y^i,y^i)),$$
The above loss $\Delta$ was used in our experiments. Thus training a structural SVM with this loss
maximizes the ROUGE-1 F score with the true manual summaries provided in the training examples 
while trading it off with  margin. Note that we could easily use a different loss function (as the method is not tied to this particualr choice) if we had a different target evaluation metric. Finally, once a ${\bf w}$ is 
obtained from the structural SVM training, a prediction summary for a test document $x$ 
can be easily obtained from \eqref{eq:svmpred}.


\section{Experiments}

In this section, we empirically evaluate the approach proposed in this
paper. 
\label{sec:expds}
Following \cite{Lin2010}, experiments were conducted on two different
datasets (DUC '03 and '04).
These datasets contain document sets with four manual summaries for
each set. For each document set, we concatenated all the articles and
split them into sentences using the tool provided with the '03
dataset.  For the supervised setting we used 10 resamplings with a
random 20/5/5 ('03) and 40/5/5 ('04) train/test/validation split. We
determining the best $C$ value using the performance on each validation
set and then report average performence over the corresponding test sets. Baseline performance (the approach of \cite{Lin2010}) was
computed using all 10 test sets as a single test set. For all
experiments and datasets, we used $r=0.3$ in the greedy algorithm as recommended in
\cite{Lin2010} for the '03 dataset. We find that changing $r$ has only a small influence on performance
\footnote{Setting $r$ to 1 and thus eliminating the non-linearity does
  lower the score (e.g. to 0.38466 for the pairwise model on DUC '03
  compared with the results on Figure \ref{fig:suppair}).}.

The construction of features for learning is organized by word groups. 
The most trivial group is simply all words (\textit{basic}). Considering
the properties of the words themselves, we constructed several features 
from properties such as capitalized words, words of certain length and non-stop words  
(\textit{cap+stop+len}). We obtained another set of features from 
the most frequently occuring words in all the articles  
 (\textit{minmax}). We also considered the position of a sentence (containing 
the word) in the article as another feature (\textit{location}).
All those word groups can then be further refined by selecting different 
thresholds, weighting schemes (e.g. TFIDF) and forming binned variants of 
these features.

For the 
pairwise model we use cosine similarity between sentences using only 
words in a given word group during computation. For the word coverage 
model we create separate features for covering words in different groups. 
This gives us fairly comparable feature strength in both models. 
The only further addition is use of different word coverage levels in
the coverage model.
First we consider how well does a sentence cover a word (e.g. a sentence
with five instances of the same word might cover it better than another with
only a single instance). And secondly we look at how important it is to
cover a word (e.g. if a word appears in a large fraction of sentences we
might want to be sure to cover it). Combining those two criteria using 
different thresholds we get a set of features for each word. Our coverage
features are motivated from the approach of \cite{r31}.
In contrast, the hand-tuned pairwise baseline uses only TFIDF weighted cosine
similarity between sentences using all words, following the approach in \cite{Lin2010}.

The resulting summaries are evaluated using ROUGE version 1.5.5 \cite{r18}. 
We selected the ROUGE-1 F measure because it was used by \cite{Lin2010} and because it is one of the commonly 
used performance scores in recent work. However, our learning method applies to other performance measures as well.
Note that we use the ROUGE-1 F measure both for the loss function during 
learning, as well as for the evaluation of the predicted summaries.

\subsection{How does learning compare to manual tuning?}

In our first experiment, we compare our supervised learning approach
to the hand-tuned approach. The results from this experiment are
summarized in Figure ~\ref{fig:suppair}. First, supervised training of
the pairwise model \cite{Lin2010} resulted in a statistically
significant ($p \le 0.05$) increase in performance on both datasets
compared to our reimplementation of the manually tuned pairwise model.
Note that our reimplementation of the approach of \cite{Lin2010}
resulted in slightly different performance numbers than those reported
in \cite{Lin2010} -- better on DUC '03 and somewhat lower on DUC '04,
if evaluated on the same selection of test examples as theirs. We
conjecture that this is due to small differences in implementation
and/or preprocessing of the dataset. Furthermore, as authors of
\cite{Lin2010} note in their paper, the '03 and '04 datasets behave
quite differently.

\begin{figure}[h]
\begin{center}
\begin{tabular}{l|l|c}
model & dataset & ROUGE-1 F (stderr) \\
\hline
pairwise & DUC '03 & \textbf{0.3929} (0.0074) \\
coverage & & \textbf{0.3784} (0.0059) \\ 
hand-tuned & & 0.3571 (0.0063) \\
\hline
pairwise & DUC '04 & \textbf{0.4066} (0.0061) \\
coverage & & 0.3992 (0.0054) \\ 
hand-tuned & & 0.3935 (0.0052) \\
\hline
\end{tabular}
\end{center}
\caption{
Results obtained on DUC '03 and '04 datasets using the supervised models.
Increase in performance over the hand-tuned is statistically significant ($p \le 0.05$) 
for the pairwise model on the both datasets, but only on DUC '03 for the coverage model. }
\label{fig:suppair}
\end{figure}

Figure ~\ref{fig:suppair} also reports the performance for the coverage model as trained by our algorithm. These results can be compared against those for the pairwise model. Since we are using features of comparable strength in both approaches, as well as the same 
greedy algorithm and structural SVM learning method, this comparison largely reflects the quality of models themselves. On the '04 dataset both models achieve 
the same performance while on '03 the pairwise model performs 
significantly ($p \le 0.05$) better than the coverage model.

Overall, pairwise model appears to perform slightly better than the
coverage model with the datasets and features we used. Therefore, we
focus on the pairwise model in the following.

\subsection{How fast does the algorithm learn?}
Hand-tuned approaches have limited flexibility. Whenever we move to a significantly 
different collection of documents we have to reinvest time to retune it. Learning can make this adaptation to a new collection more automatic and faster -- especially since training data has to be collected even for manual tuning.

\begin{figure}
\begin{center}
\includegraphics[width=80mm]{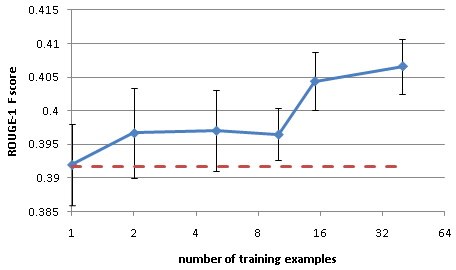}
\end{center}
\caption{
Learning curve for the pairwise model on DUC '04 dataset showing ROUGE-1 F scores for different numbers of
learning examples (logarithmic scale). The dashed line represents the preformance of the hand-tuned model.}
\label{fig:lcrv}
\end{figure}

Figure \ref{fig:lcrv} evaluates how effectively the learning algorithm can make use of a given amount of training data. In particular, the figure shows the learning curve for our approach. Even with very few training examples the learning approach already outperforms the baseline. Furthermore, at the maximum number of training examples available to us the curve still increases. We therefore conjecture that more data would further improve performance.

\subsection{Where is room for improvement?}

To get a rough estimate of what is actually achievable in terms of the final ROUGE-1 F score
we looked at different ``upper bounds'' under various scenarios (Figure \ref{fig:ubound}).
First, ROUGE score is computed by using four manual summaries from different
assessors, so that we can estimate inter-subject disagreement. If one computes the ROUGE score of a held-out summary against the remaining three summaries, the resulting performance is given in the row
\textit{human} of Figure \ref{fig:ubound}. It provides a reasonable estimate of human performance. 

Second, in extractive summarization we restrict summaries to sentences from the documents themselves, which is likely to lead to a reduction in ROUGE. To estimate this drop, we use the greedy algorithm to select the extractive summary that maximizes ROUGE on the test documents. The resulting performance is given in the row \textit{extractive} of Figure \ref{fig:ubound}. On both dataset, the drop in performance for this (approximately\footnote{We compared the greedy algorithm with exhaustive search for up to three selected sentences 
(more than that would take too long). In about half the cases we got the same solution, in other
cases the soultion was on average about 1\% below optimal confirming that greedy selection
works quite well.}) optimal extractive summary is about 10 points of ROUGE.

Third, we expect some drop in performance, since our model may not be able to fit the optimal extractive summaries due to a lack of expressiveness. This can be estimated by looking at training set performance, as reported in row \textit{model fit} of Figure \ref{fig:ubound}. On both datasets, we see a drop of about 5 points of ROUGE performance. Adding more and better features
might help the model fit the data better. 

Finally, a last drop in performance may come from overfitting. The test set ROUGE scores are given in the row \textit{prediction} of Figure \ref{fig:ubound}. Note that the drop between training and test performance is rather small, so overfitting is not an issue and is well controlled in our algorithm. We therefore conclude that increasing model fidelity seems like a promising direction for further improvements.

\begin{figure}[h]
\begin{center}
\begin{tabular}{l|l|c}
bound & dataset & ROUGE-1 F \\
\hline
human & DUC '03 & 0.56235 \\
extractive & & 0.45497 \\
model fit & & 0.40873 \\
prediction & & 0.39294 \\
\hline
human & DUC '04 & 0.55221 \\
extractive & & 0.45199 \\
model fit & & 0.40963 \\
prediction & & 0.40662 \\
\hline
\end{tabular}
\end{center}
\caption{
Upper bounds on ROUGE-1 F scores: agreement between manual summaries, 
greedily computed best extractive summaries, best model fit on the train set
(using the best $C$ value) and the test scores of the pairwise model. }
\label{fig:ubound}
\end{figure}

\subsection{Which features are most useful?}
To understand which features affected the final performance of our approach,
we assessed the strength of each set of our features. In particular, we 
looked at how the final test score changes when we removed certain features
groups (described in the beginning of Section \ref{sec:expds}) as shown in Figure \ref{fig:feats}.

The most important group of features are the basic features (pure cosine
similarity between sentences) since removing them results in the largest drop in performance.
However, other features play a significant role too (i.e. only
the basic ones are not enough to achieve good performance). This confirms that 
performance can be improved by adding richer fatures 
instead of using only a single similarity score as in \cite{Lin2010}. Using learning for these complex model is essential, since hand-tuning is likely to be intractable. 

The second most important group 
of features considering the drop in performance (i.e. \textit{location}) looks at 
positions of sentences in the articles. This makes intuitive sense because the first 
sentences in news articles is usually packed with informatin. The other three groups do
not have a significant impact on their own.

\begin{figure}[h]
\begin{center}
\begin{tabular}{l|c}
removed & ROUGE-1 F \\
group & \\
\hline
\hline
none & 0.40662 \\
\hline
basic & \textbf{0.38681} \\
all except basic & \textbf{0.39723} \\
location & \textbf{0.39782} \\
sent+doc & 0.39901 \\
cap+stop+len & 0.40273 \\
minmax & 0.40721 \\
\hline
\end{tabular}
\end{center}
\caption{
Effects of removing different feature groups on the DUC '04 dataset.
Bold font marks significant difference ($p \le 0.05$) when compared to the full pariwise model.
The most important are basic similarity features including all words
(similar to \cite{Lin2010}). The last feature group actually lowered the score
but is included in the model because we only found this out later on DUC '04 dataset.}
\label{fig:feats}
\end{figure}

\subsection{How important is it to train with multiple summaries?}

While having four manual summaries may be important for computing a reliable ROUGE score for evaluation, it is not clear whether such an approach is the most efficient use of annotator resources for training. In our final experiment, we trained our method using only a single manual summary for each set of documents. 
When using only a single manual summary, we arbitrarily took the first one out of the provided four
reference summaries and used only it to compute the target label for training
(instead of using average loss towards all four of them). Otherwise, the experimental setup was the same as in the previous subsections, using the pairwise model.

For DUC '04, the ROUGE-1 F score obtained using only a single summary per 
document set was 0.4010, which is slightly but not significantly lower than the 0.4066 obtained with four summaries (as shown on Figure \ref{fig:suppair}). Similarly, on DUC '03 the performance drop from 0.3929 to 0.3838 was not significant as well.

Based on those results, we conjecture that having more documents sets with only a single 
manual summary is more useful for training than fewer training examples with better labels
(i.e. multiple summaries). In both cases, we spend approximately the same amount of effort
(as the summaries are the most expensive component of the training data), however having more
training examples helps (according to the learning curve presented before) while spending 
effort on multiple summaries appears to have only minor benefit for training.


\section{Conclusions}

This paper presented a supervised learning approach to extractive document summarization based on structual SVMs. The learning method applies to all submodular scoring functions, ranging from pairwise-similarity models to coverage-based approaches. The learning problem is formulated into a convex quadratic program and then solved approximated using a cutting-plane method. In an empirical evaluation, the structural SVM approach  
significantly outperforms conventional hand-tuned models on the 
DUC '03 and '04 datasets. A key advantage of the learning approach is its ability to handle large numbers of features, providing substantial flexibility for building high-fidelity summarization models. Furthermore, it shows good control of overfitting, making it possible to train models even with only a few training examples.


\section{Acknowledgments}

We thank Claire Cardie and the members of the Cornell NLP Seminar for their valuable feedback. This research was funded in part through NSF Awards IIS-0812091 and IIS-0905467.


\end{document}